\documentclass[11pt,a4paper]{article}
\usepackage[hyperref]{eacl2021}
\usepackage{times}
\usepackage{latexsym}

\definecolor{babyblue}{rgb}{0.54, 0.81, 0.94}
\definecolor{harvardcrimson}{rgb}{0.79, 0.0, 0.09}
\definecolor{hansayellow}{rgb}{0.91, 0.84, 0.42}
\definecolor{docgrey}{rgb}{0.9, 0.9, 0.9}
\definecolor{darkgreen}{rgb}{0.13725490196078433, 0.4196078431372549, 0.03529411764705882}
\definecolor{darkgred}{rgb}{0.5882352941176471, 0.11764705882352941, 0.047058823529411764}
\definecolor{darkblue}{rgb}{0.0392156862745098, 0.12941176470588237, 0.5411764705882353}

\usepackage{microtype}

\usepackage[boldmath]{numprint}
\usepackage{booktabs}
\usepackage{multirow}
\usepackage{graphicx}
\usepackage{xcolor}
\usepackage{csquotes}
\usepackage{amsmath}
\usepackage{amssymb}
\usepackage{siunitx}
\usepackage{hyperref}

\newcommand\redbf[1]{\textcolor{darkgred}{\textbf{#1}}}
\newcommand\bluebf[1]{\textcolor{darkblue}{\textbf{#1}}}
\newcommand\greenbf[1]{\textcolor{darkgreen}{\textbf{#1}}}

\renewcommand{\bfseries}{\fontseries{b}\selectfont}
\newrobustcmd{\B}{\bfseries}

\newcommand\name{JEREX}

\aclfinalcopy 

\title{An End-to-end Model for Entity-level Relation Extraction \\ using Multi-instance Learning}

\author{Markus Eberts \hspace{1cm} Adrian Ulges\\
  RheinMain University of Applied Sciences \\
  Wiesbaden, Germany \\
  \texttt{\{markus.eberts, adrian.ulges\}@hs-rm.de}}

\date{}

\begin{document}
\maketitle
\begin{abstract}
We present a joint model for entity-level relation extraction from documents. In contrast to other approaches -- which focus on local intra-sentence mention pairs and thus require annotations on mention level -- our model operates on entity level. To do so, a multi-task approach is followed that builds upon coreference resolution and gathers relevant signals via multi-instance learning with multi-level representations combining global entity and local mention information. 
We achieve state-of-the-art relation extraction results on the DocRED dataset and report the first entity-level end-to-end relation extraction results for future reference.
Finally, our experimental results suggest that a joint approach is on par with task-specific learning, though more efficient due to shared parameters and training steps.
\end{abstract}

\section{Introduction}
Information extraction addresses the inference of formal knowledge (typically, entities and relations) from text. The field has recently experienced a significant boost due to the development of neural approaches~\cite{zeng:2014:rel_cnn, zhang:2015:rel_pos, kumar:2017:rel_survey}. This has led to two shifts in research: First, while earlier work has focused on sentence level relation extraction~\cite{hendrickx:2010:semeval,han:2018:fewrel,zhang:2017:tacred}, more recent models extract facts from longer text passages (document-level). This enables the detection of inter-sentence relations that may only be implicitly expressed and require reasoning across sentence boundaries. Current models in this area do not rely on mention-level annotations and aggregate signals from multiple mentions of the same entity. 

The second shift has been towards multi-task learning: While earlier approaches tackle entity mention detection and relation extraction with separate models, recent joint models address these tasks at once~\cite{bekoulis:2018:multi_head,nguyen:2019:biaffine_attention,wadden:2019:dygie++}. This does not only improve simplicity and efficiency, but is also commonly motivated by the fact that tasks can benefit from each other: For example, knowledge of two entities' types (such as {\it person}+{\it organization}) can boost certain relations between them (such as {\it ceo\_of}). 

\begin{figure} 
\centering
\framebox{
\begin{tabular}{ p{7cm} }
    The \bluebf{Portland Golf Club} is a private golf club in the northwest \redbf{United States}, in suburban Portland, Oregon. The \bluebf{PGC} is located in the unincorporated \greenbf{Raleigh Hills} area of eastern Washington County, southwest of downtown Portland and east of Beaverton. \bluebf{PGC} was established in the winter of \textbf{1914}, when a group of nine businessmen assembled to form a new club after leaving their respective clubs. The \bluebf{golf club} hosted the Ryder Cup matches of 1947, the first renewal in a decade, due to World War II. The \redbf{U.S.} team defeated Great Britain 11 to 1 in wet conditions in early November.
\end{tabular}
}
\caption{Our goal is to perform end-to-end entity-level relation extraction on whole documents. We extract entity mentions (\enquote{PGC}), entity clusters (\{Portland Golf Club, PGC, golf club\}), their types ($ORG$) and relations to other entities in the document, such as (\{Portland Golf Club, PGC, golf club\}$_{ORG}$, \emph{inception}, \{1914\}$_{TIME}$), with a single, joint model. Note that document-level relation extraction requires the aggregation of relevant information from multiple sentences, such as in (\{Raleigh Hills\}$_{LOC}$, \emph{country}, \{United States, U.S.\})$_{LOC}$). Other entities in the example document are omitted for clarity.}
\label{fig:doc_example} 
\end{figure}

We follow this line of research, and present \name{}\footnote{The code for reproducing our results is available at \href{https://github.com/lavis-nlp/jerex}{https://github.com/lavis-nlp/jerex}.} (\enquote{\textbf{J}oint \textbf{E}ntity-Level \textbf{R}elation \textbf{Ex}tractor}), a novel approach for joint information extraction. 
\name{} is to our knowledge the first approach that combines a multi-task model with entity-level relation extraction: In contrast to previous work, our model jointly learns relations and entities without annotations on mention level, but extracts document-level entity clusters and predicts relations between those clusters using a \emph{multi-instance learning} (MIL)~\cite{dietterich:1997:mil, riedel:2010:nyt, surdeanu:2012:mil_multi_label} approach.
The model is trained jointly on mention detection, coreference resolution, entity classification and relation extraction (Figure~\ref{fig:doc_example}).

While we follow best practices for the first three tasks, we propose a novel representation for relation extraction, which combines global entity-level representations with localized mention-level ones. 
We present experiments on the DocRED~\cite{yao:2019:docred} dataset for entity-level relation extraction. Though it is arguably simpler compared to recent graph propagation models~\cite{nan:2020:bert_lsr} or special pre-training~\cite{ye:2020:coref_bert}, our approach achieves state-of-the-art results. 
 
We also report the first results for end-to-end relation extraction on DocRED as a reference for future work. In ablation studies we show that (1) combining a global and local representations is beneficial, and (2) that joint training appears to be on par with separate per-task models.

\section{Related Work}
Relation extraction is one of the most studied natural language processing (NLP) problems to date. Most approaches focus on classifying the relation between a given entity mention pair. Here various neural network based models, such as RNNs~\cite{zhang:2015:rel_pos}, CNNs~\cite{zeng:2014:rel_cnn}, recursive neural networks~\cite{socher:2012:mv_rnn} or Transformer-type architectures~\cite{wu:2019:semeval_bert} have been investigated. However, these approaches are usually limited to local, intra-sentence, relations and are not suited for document-level, inter-sentence, classification. Since complex relations require the aggregation of information distributed over multiple sentences, document-level relation extraction has recently drawn attention (e.g. ~\citealt{quirk:2017:distant_intra_sentence_re,verga:2018:multi_instance,gupta:2019:intra_inter_re,yao:2019:docred}). Still, these models rely on specific entity mentions to be given. While progress in the joint detection of entity mentions and intra-sentence relations has been made~\cite{gupta:2016:table_filling, bekoulis:2018:multi_head, luan:2018:scierc}, the combination of coreference resolution with relation extraction for entity-level reasoning in a single, jointly-trained, model is widely unexplored. 

\paragraph{Document-level Relation Extraction} Recent work on document-level relation extraction directly learns relations between entities (i.e. clusters of mentions referring to the same entity) within a document, requiring no relation annotations on mention level. To gather relevant information across sentence boundaries, multi-instance learning has successfully been applied to this task. In multi-instance learning, the goal is to assign labels to bags (here, entity pairs), each containing multiple instances (here, specific mention pairs). \citet{verga:2018:multi_instance} apply multi-instance learning to detect domain-specific relations in biological text. They compute relation scores for each mention pair of two entity clusters and aggregate these scores using a smooth max-pooling operation. ~\citet{christopoulou:2019:dots} and~\citet{sahu:2019:multi_instance_graph} improve upon \citet{verga:2018:multi_instance} by constructing document-level graphs to model global interactions. 
While the aforementioned models tackle very specific domains with few relation types, the recently released DocRED dataset~\cite{yao:2019:docred} enables general-domain research on a rich relation type set (96 types). \citet{yao:2019:docred} provide several baseline architectures, such as CNN-, LSTM- or Transformer-based models, that operate on global, mention averaged, entity representations. \citet{wang:2019:two-step-bert} use a two-step process by identifying related entities in a first step and classifying them in a second step.~\citet{Tang:2020:hin} employ a hierarchical inference network, combining entity representations with attention over individual sentences to form the final decision.
~\citet{nan:2020:bert_lsr} apply a graph neural network ~\cite{kipf:2017:gcn} to construct a document-level graph of mention, entity and meta-dependency nodes. 
The current state-of-the-art constitutes the CorefRoBERTa model proposed by \citet{ye:2020:coref_bert}, a RoBERTa~\cite{liu:2019:roberta} variant that is pre-trained on detecting co-referring phrases. They show that replacing RoBERTa with CorefRoBERTa improves performance on DocRED.

All these models have in common that entities and their mentions are both assumed to be given. In contrast, our approach extracts mentions, clusters them to entities, and classifies relations jointly. 

\newpage
\paragraph{Joint Entity Mention and Relation Extraction}
Prior joint models focus on the extraction of mention-level relations in sentences. Here, most approaches detect mentions by BIO (or BILOU) tagging and pair detected mentions for relation classification, e.g.~\cite{gupta:2016:table_filling, zhou:2017:joint_hybrid, zheng:2017:joint_novel_tagging, bekoulis:2018:multi_head, nguyen:2019:biaffine_attention, miwa:2016:stacked_rnn}. However, these models are not able to detect relations between overlapping entity mentions. Recently, so-called span-based approaches~\cite{lee:2017:span_coreference} were successfully applied to this task~\cite{luan:2018:scierc, eberts:2020:spert}: By enumerating each token span of a sentence, these models handle overlapping mentions by design. \citet{wolf:2018:hierarch_multi_task} train a multi-task model on named entity recognition, coreference resolution and relation extraction. By adding coreference resolution as an auxilary task, ~\citet{luan:2019:span_graphs} propagate information through coreference chains. Still, these models rely on mention-level annotations and only detect intra-sentence relations between mentions, whereas our model explicitly constructs clusters of co-referring mentions and uses these clusters to detect complex entity-level relations in long documents using multi-instance reasoning.

\section{Approach} \label{sec:approach}
\begin{figure*}[ht]
    \centering
    \includegraphics[width=1.\textwidth]{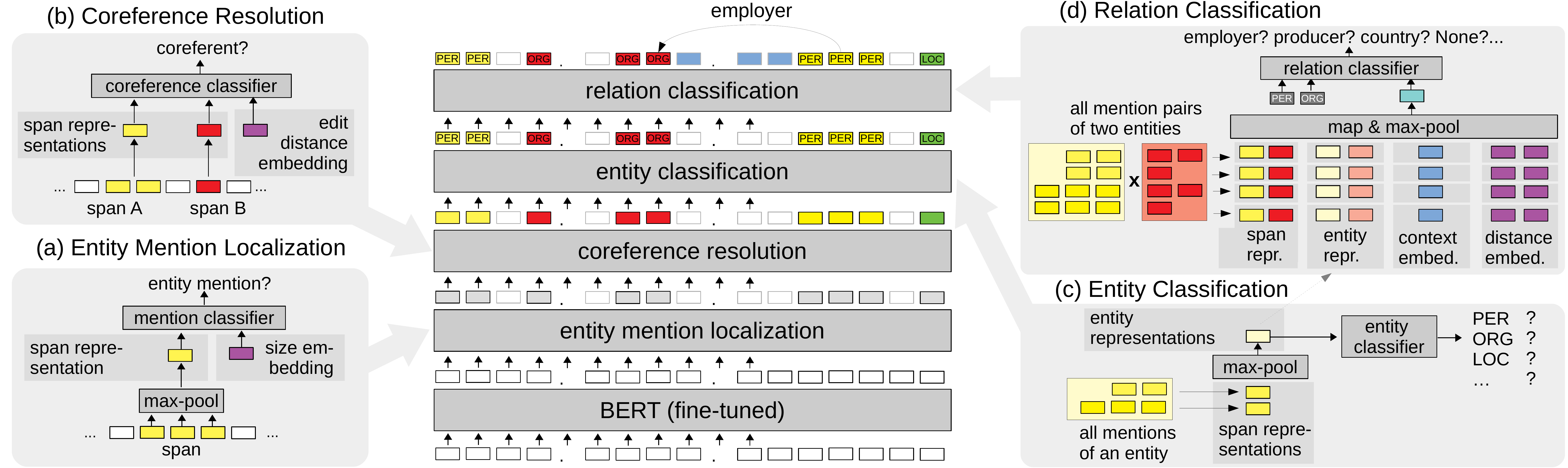}
    \caption{Our approach combines entity mention localization (a), coreference resolution (b), entity classification (c) and relation classification (d) within a joint multi-task model, which is trained jointly on entity-level relation extraction. The sub-components share a single BERT encoder for document encoding. Each input document is only encoded once (\emph{single-pass}) to speed-up training/inference, with sub-components operating on the contextualized embeddings. Both entity classification and relation classification use multi-instance learning to synthesize relevant signals scattered throughout the input document.}
    \label{fig:approach}
\end{figure*}

\name{} processes documents containing multiple sentences and extracts entity mentions, clusters them to entities, and outputs types and relations on entity level. \name{} consists of four task-specific components, which are based on the same encoder and mention representations, and are trained in a joint manner. An input document is first tokenized, yielding a sequence of $n$  byte-pair encoded (BPE) \cite{sennrich:2016:bpe} tokens. We then use the pre-trained Transformer-type network BERT \cite{devlin:2018:bert} to obtain a contextualized embedding sequence $(\mathbf{e}_1, \mathbf{e}_2, ... \mathbf{e}_n)$ of the document. Since our goal is to perform end-to-end relation extraction, neither entities nor their corresponding mentions in the document are known in inference. 

\subsection{Model Architecture}
We suggest a multi-level model: First, we localize all entity mentions in the document (a) by a \emph{span-based} approach~\cite{lee:2017:span_coreference}. After this, 
detected mentions are clustered into entities by \emph{coreference resolution} (b). We then classify the type (such as \emph{person} or \emph{company}) of each entity cluster by a fusion over local mention representations (\emph{entity classification}) (c). Finally, relations between entities are extracted by a reasoning over mention pairs (d). The full model architecture is illustrated in Figure~\ref{fig:approach}.

\paragraph{(a) Entity Mention Localization} Here our model performs a search over all document token subsequences (or \emph{spans}). In contrast to BIO/BILOU-based approaches for entity mention localization, span-based approaches are able to detect overlapping mentions. Let $s := (\mathbf{e}_i, \mathbf{e}_{i+1},$ $ ..., \mathbf{e}_{i+k})$ denote an arbitrary candidate span. Following~\citet{eberts:2020:spert}, we first obtain a span representation by max-pooling the span's token embeddings: 
\begin{equation}
\label{eq:spanrepr}
\mathbf{e}(s) := \text{max-pool}(\mathbf{e}_i, \mathbf{e}_{i+1}, ..., \mathbf{e}_{i+k})
\end{equation}
Our \emph{mention classifier} takes the span representation $\mathbf{e}(s)$ as well as a span size embedding $\mathbf{w}_{k+1}^s$ \cite{lee:2017:span_coreference} as meta information. We perform binary classification and use a sigmoid activation to obtain a probability for $s$ to constitute an entity mention: 
\begin{equation}
\label{eq:spanclassifier}
\hat{y}^s = \sigma \Big(\text{FFNN}^s(\mathbf{e}(s) \circ \mathbf{w}_{k+1}^s) \Big)
\end{equation}
where $\circ$ denotes concatenation and $\text{FFNN}^s$ is a two-layer feedforward network with an inner ReLu activation.
Span classification is carried out on all token spans up to a fixed length $L$. We apply a filter threshold $\alpha^s$ on the confidence scores, retaining all spans with $\hat{y}^s	\geq \alpha_s$ and  leaving a set $\mathcal{S}$ of spans supposedly constituting entity mentions.

\paragraph{(b) Coreference Resolution} Entity mentions referring to the same entity (e.g. \enquote{Elizabeth II.} and \enquote{the Queen}) can be scattered throughout the input document. To later extract relations on entity level, local mentions need to be grouped to document-level entity clusters by coreference resolution. We use a simple mention-pair~\cite{soon:2001:coref_mention_pair} model: Our component classifies pairs $(s_1,s_2) \in \mathcal{S} {\times} \mathcal{S}$ of detected entity mentions as coreferent or not, by combining
the span representations $\mathbf{e}(s_1)$ and $\mathbf{e}(s_2)$ with an edit distance embedding $\mathbf{w}_{d}^c$: We compute the Levenshtein distance~\cite{levenshtein:1966:levenshtein} between spans $d := D(s_1,s_2)$ and use a learned embedding $\mathbf{w}_{d}^c$.
A mention pair representation $\mathbf{x}^c$ is constructed by concatenation:
\begin{equation}
\label{eq:corefrepr}
\mathbf{x}^c := \mathbf{e}(s_1) \circ \mathbf{e}(s_2) \circ \mathbf{w}_{d}^c
\end{equation}
Similar to span classification, we conduct binary classification using a sigmoid activation, obtaining a similarity score between the two mentions: 
\begin{equation}
\label{eq:corefscore}
\hat{y}^c := \sigma \Big( \text{FFNN}^c(\mathbf{x}^c) \Big)
\end{equation}
where $\text{FFNN}^c$ follows the same architecture as $\text{FFNN}^s$.
We construct a similarity matrix $C \in \mathbb{R}^{m \times m}$ (with $m$ referring to the document's overall number of mentions) containing the similarity scores between every mention pair. By applying a filter threshold $\alpha^c$, we cluster mentions using complete linkage~\cite{muellner:2011:clustering}, yielding a set $\mathcal{E}$ containing clusters of entity mentions. We refer to these clusters as \emph{entities} or \emph{entity clusters} in the following.

\paragraph{(c) Entity Classification} Next, we map each entity to a type such as $location$ or $person$: We first fuse the mention representations of an entity cluster $\{s_1, s_2, ..., s_t\} \in \mathcal{E}$ by max-pooling:
\begin{equation} 
\label{eq:entityrepr}
\mathbf{x}^e := \text{max-pool}(\mathbf{e}(s_1), \mathbf{e}(s_2), ..., \mathbf{e}(s_t))
\end{equation}
Entity classification is then carried out on the entity representation $\mathbf{x}^e$, allowing the model to draw information from mentions spread across different parts of the document. $\mathbf{x}^e$ is fed into a softmax classifier, yielding a probability distribution over the entity types:
\begin{equation}
\label{eq:corefclassifier}
\hat{y}^e := \text{softmax} \Big( \text{FFNN}^e(\mathbf{x}^e) \Big)
\end{equation}
We assign the highest scored type to the entity.

\paragraph{(d) Relation Classification}
Our final component assigns relation types to pairs of entities.
Note that the directionality, i.e. which entity constitutes the head/tail of the relation, needs to be inferred, and that the input document can express multiple relations between different mentions of the same entity pair. Let $\mathcal{R}$ denote a set of pre-defined relation types. The relation classifier processes each entity pair $(e_1, e_2) \in \mathcal{E} {\times} \mathcal{E}$, estimating which, if any, relations from $\mathcal{R}$ are expressed between these entities. To do so, we score every candidate triple ($e_1{,} r_i{,} e_2$), 
expressing that $e_1$ (as head) is in relation $r_i$ with $e_2$ (as tail).  We design two types of relation classifiers: A \emph{global relation classifier}, serving as a baseline, which consumes the entity cluster representations $\mathbf{x}^e$, and a \emph{multi-instance classifier}, which assumes that certain entity mention pairs support specific relations and synthesizes this information into an entity-pair level representation.

\paragraph{Global Relation Classifier (GRC)} The global classifier builds upon the max-pooled entity cluster representations $\mathbf{x}_1^e$ and $\mathbf{x}_2^e$ of an entity pair $(e_1, e_2)$. We further embed the corresponding entity types ($\mathbf{w}_1^e$ / $\mathbf{w}_2^e$), which was shown to be beneficial in prior work~\cite{yao:2019:docred}, and compute an entity-pair representation by concatenation:
\begin{equation} \label{eq:entitypairrepr}
\mathbf{x}^p := \Big( \mathbf{x}_1^e \circ \mathbf{w}_1^e \Big) \circ \Big( \mathbf{x}_2^e \circ \mathbf{w}_2^e \Big)
\end{equation}

This representation is fed into a 2-layer FFNN (similar to FFNN$^s$), mapping it to the number of relation types $\#\mathcal{R}$. The final layer features sigmoid activations for multi-label classification and assigns any relation type exceeding a threshold $\alpha^r$:

\begin{equation}
\label{eq:globalrelclassifier}    \hat{y}^r := \sigma \Big( \text{FFNN}^p(\mathbf{x}^p) \Big)
\end{equation}

\paragraph{Multi-instance Relation Classifier (MRC)} In contrast to the global classifier (GRC), the multi-instance relation classifier operates on mention level: Since only entity-level labels are available, we treat entity mention pairs as latent variables and estimate relations by a fusion over these mention pairs. For any pair of entity clusters $e_1{=}\{s_1^1, s_2^1, ..., s_{t_1}^1\}$ and $e_2{=}\{s_1^2, s_2^2, ..., s_{t_2}^2\}$, we compute a mention-pair representation for any $(s_1, s_2) {\in} e_1 {\times} e_2$. This representation is obtained by concatenating the global entity embeddings (Equation \eqref{eq:entityrepr}) with the mentions' local span representations (Equation \eqref{eq:spanrepr})
\begin{equation}
\label{eq:mentioninput1}
\mathbf{u}(s_1,s_2) := \Big( \mathbf{e}(s_1) \circ \mathbf{x}^e_1 \Big) \circ \Big(  \mathbf{e}(s_2) \circ \mathbf{x}^e_2 \Big)
\end{equation}
Further, as we expect close-by mentions to be stronger indicators of relations, we add meta embeddings for the {\it distances} $d_s$,$d_t$ between the two mentions, both in sentences ($d_s$) and in tokens ($d_t$). In addition, following \citet{eberts:2020:spert}, the max-pooled context between the two mentions ($\mathbf{c}(s_1, s_2)$) is added. This \emph{localized context} provides a more focused view on the document and was found to be especially beneficial for long, and therefore noisy, inputs:
\begin{equation}
\label{eq:mentioninput2}
\mathbf{u'}(s_1{,}s_2) {:=} \mathbf{u}(s_1{,}s_2) \circ \mathbf{c}(s_1{,} s_2) \circ \mathbf{w}_{d_s}^r \circ 
\mathbf{w}_{d_t}^{r'}
\end{equation}
This mention-pair representation is mapped by a single feed-forward layer to the original token embedding size ($768$):
\begin{equation}
\label{eq:mentionpairrepr}
\mathbf{u''}(s_1, s_2) := \text{FFNN}^p(\mathbf{u'}(s_1, s_2))
\end{equation}
These focused representations are then combined by max-pooling: 
\begin{equation} \label{eq:fusedrepr}
\mathbf{x}^r {=} \text{max-pool}(\{\mathbf{u''}(s_1, s_2) | s_1{\in}e_1{,} s_2{\in} e_2\})
\end{equation}
Akin to GRC, we concatenate $\mathbf{x}^r$ with entity type embeddings $\mathbf{w}_1^e/\mathbf{w}_2^e$ and apply a two-layer FFNN (again, similar to FFNN$^s$).
Note that for both classifiers (GRC/MRC), we need to score both ($s_1$, $r_i$, $s_2$) and ($s_2$, $r_i$, $s_1$) to infer the direction of asymmetric relations.

\subsection{Training} \label{sec:training}

\npdecimalsign{.}
\nprounddigits{2}
\begin{table*}
\sisetup{round-mode=places,detect-weight}
\centering
\begin{tabular}{c l S S S S S S}
\toprule
      & & \multicolumn{3}{c}{{\textbf{Joint Model}$^*$}} & \multicolumn{3}{c}{{\textbf{Pipeline}}} \\ \cmidrule(lr){3-5} \cmidrule(lr){6-8}
     \textbf{Level} & \textbf{Task} & {Precision} & {Recall} & {F1} & {Precision}&{Recall}&{F1} \\ \midrule
     (a) & Mention Localization & 93.28913423754554 & 92.70222023875934 & 92.99438038342063 & 92.87034924830034 & 92.45788240544462 & 92.66362618180385 \\
     (b) & Coreference Resolution & 82.51984790001059 & 83.05892738909733 & 82.78786712808218 & 82.11101710467351 & 82.66166409181197 & 82.38517453455762 \\
     (c) & Entity Classification & 79.8436351712385 & 80.35898190378107 & 80.09985709880569 & 78.9993913237506 & 79.52331911137267 & 79.2602527755901 \\
     \multirow{2}{*}{(d)} & Relation Classification & 42.75927494795886 & 38.247410947991355 & 40.375261254251676 & 43.606748210084604 & 37.50312962330716 & 40.32059813379855 \\
      & Relation Classification (GRC) & 38.68889837573745 & 37.31876635939456 & 37.9813724906379 & 39.06633073375847 & 36.43564356435643 & 37.69679991072489 \\
     \bottomrule
\end{tabular}
\caption{Test set evaluation results of our multi-level end-to-end system \name{} on DocRED (using the end-to-end split). We either train the model jointly on all four sub-components (left) or arrange separately trained models in a pipeline (right) 
($^*$ joint results are for MRC except for the last row).} 
\label{table:joint_results} 
\end{table*}

We perform a supervised multi-task training, whereas each training document features ground truth for all four subtasks (mention localization, coreference resolution, as well as entity and relation classification). We optimize the joint loss of all four components:
\begin{equation}
\mathcal{L} := \beta_s \cdot \mathcal{L}^s + \beta_c \cdot \mathcal{L}^c + \beta_e \cdot \mathcal{L}^e + \beta_r \cdot  \mathcal{L}^r
\end{equation}
$\mathcal{L}^s$, $\mathcal{L}^c$ and $\mathcal{L}^r$ denote the binary cross entropy losses of the span, coreference and relation classifiers. We use a cross entropy loss ($\mathcal{L}^e$) for the entity classifier. A batch is formed by drawing positive and negative samples from a single document for all components. We found such a {\it single-pass approach} to offer significant speed-ups both in learning and inference:
\begin{itemize}
    \item Entity mention localization: We utilize all ground truth entity mentions $\mathcal{S}^{gt}$ of a document as positive training samples, and sample a fixed number $N_s$ of random non-mention spans up to a pre-defined length $L_s$ as negative samples. Note that we only train and evaluate on the full tokens according to the dataset's tokenization, i.e. not on byte-pair encoded tokens, to limit computational complexity. Also, we only sample intra-sentence spans as negative samples. Since we found intra-mention spans to be especially challenging (\enquote{New York} versus \enquote{New York City}), we sample up to $\frac{N_s}{2}$ intra-mention spans as negative samples.
    \item Coreference resolution: The coreference classifier is trained on all span pairs drawn from ground truth entity clusters $\mathcal{E}^{gt}$ as positive samples. We further sample a fixed number $N_c$ of pairs of random ground truth entity mentions that do not belong to the same cluster as negative samples. 
    \item Entity classification: Since the entity classifier only receives clusters that supposedly constitute an entity during inference, it is trained on all ground truth entity clusters of a document.
    \item Relation classification: Here we use ground truth relations between entity clusters as positive samples and $N_r$ negative samples drawn from $\mathcal{E}^{gt} {\times} \mathcal{E}^{gt}$ that are unrelated according to the ground truth.
\end{itemize}

Each component's loss is obtained by averaging over all samples. We learn the weights and biases of sub-component specific layers as well as the meta embeddings during training. BERT is fine-tuned in the process.

\section{Experiments} \label{sec:experiments}

We evaluate \name{} on the DocRED dataset~\cite{yao:2019:docred}. DocRED ist the most diverse relation extraction dataset to date (6 entity and 96 relation types). It includes over 5,000 documents, each consisting of multiple sentences. According to~\citet{yao:2019:docred}, DocRED requires multiple types of reasoning, such as logical or common-sense reasoning, to infer relations. 

Note that previous work only uses DocRED for relation extraction (which equals our relation classifier component) and assumes entities to be given (e.g.~\citealt{wang:2019:two-step-bert, nan:2020:bert_lsr}). On the other hand, DocRED is exhaustively annotated with mentions, entities and entity-level relations, making it suitable for end-to-end systems. 
Therefore, we evaluate \name{} both as a relation classifier (to compare it with the state-of-the-art) and as a joint model (as reference for future work on joint entity-level relation extraction).  

While prior joint models focus on mention-level relations (e.g.~\citealt{gupta:2016:table_filling, bekoulis:2018:multi_head, chi:2019:hierarch_attention}), we extend the strict evaluation setting to entity level: A mention is counted as correct if its span matches a ground truth mention span. An entity cluster is considered correct if it matches the ground truth cluster exactly and the corresponding mention spans are correct. Likewise, an entity is considered correct if the cluster as well as the entity type matches a ground truth entity. Lastly, we count a relation as correct if its argument entities as well as the relation type are correct. We measure precision, recall and micro-F1 for each sub-task and report micro-averaged scores.

\paragraph{Dataset split} The original DocRED dataset is split into a train (3,053 documents), dev (1,000) and test (1,000) set. However, test relation labels are hidden and evaluation requires the submission of results via Codalab. To evaluate end-to-end systems, we form a new split by merging train and dev. We randomly sample a train (3,008 documents), dev (300 documents) and test set (700 documents). Note that we removed 45 documents since they contained wrongly annotated entities with mentions of different types. Table~\ref{table:joint_split} contains statistics of our end-to-end split\footnote{Note that DocRED contains some duplicate annotations. These are included in the statistics, but are filtered for evaluation in the end-to-end setting.}. We release the split as a reference for future work.

\npdecimalsign{.}
\nprounddigits{2}
\begin{table}
\sisetup{round-mode=places,detect-weight}
\centering
\begin{tabular}{l c c c c }
\toprule
    \textbf{Split} & {\#Doc.} & {\#Men.} & {\#Ent.} & {\#Rel.} \\ \midrule
     Train & 3,008 & 78,677 & 58,708 & 37,486 \\
     Dev & 300 & 7,702 & 5,805 & 3,678 \\
     Test & 700 & 17,988 & 13,594 & 8,787 \\
     Total & 4,008 & 104,367 & 78,107 & 49,951 \\
     \bottomrule
\end{tabular}
\caption{DocRED dataset split used for end-to-end relation extraction.} 
\label{table:joint_split} 
\end{table}

\paragraph{Hyperparameters} We use $\text{BERT}_{\text{BASE}}$ (cased)\footnote{We use the implementation from~\cite{wolf:2019:hugging_face}.} for document encoding, an attention-based language model pre-trained on English text~\cite{devlin:2018:bert}. Hyperparameters were tuned on the end-to-end dev set: We adopt several settings from~\cite{devlin:2018:bert}, including the usage of the Adam Optimizer with a linear warmup and linear decay learning rate schedule, a peak learning rate of 5e-5\footnote{We performed a grid search over [5e-6, 1e-5, 5e-5, 1e-4, 5e-4].} and application of dropout with a rate of $0.1$ throughout the model. We set the size of meta embeddings ($\mathbf{w}^s$, $\mathbf{w}^c$, $\mathbf{w}^e$, $\mathbf{w}_{d_s}^r$, $\mathbf{w}_{d_t}^{r'}$) to $25$ and the number of epochs to $20$. Performance is measured once per epoch on the dev set, out of which the best performing model is used for the final evaluation on the test set. A grid search is performed for the mention, coreference and relation filter threshold ($\alpha^s{=}0.85$, $\alpha^c{=}0.85$, $\alpha^r (\text{GRC}){=}0.55$, $\alpha^r (\text{MRC}){=}0.6$) with a step size of 0.05. The number of negative samples ($N_s{=}N_c{=}N_r{=}200$) and sub-task loss weights ($\beta_s{=}\beta_c{=}\beta_r{=}1$, $\beta_e{=}0.25$) are manually tuned. Note that some documents in DocRED exceed the maximum context size of BERT ($512$ BPE tokens). In this case we train the remaining position embeddings from scratch.

\subsection{End-to-End Relation Extraction}

\npdecimalsign{.}
\nprounddigits{2}
\begin{table}
\sisetup{round-mode=places,detect-weight}
\centering
\begin{tabular}{l S S }
\toprule
    \textbf{Model} & {Ign F1} & {F1} \\ \midrule
    CNN~\cite{yao:2019:docred} & 40.33 & 42.26  \\
    LSTM~\cite{yao:2019:docred} & 47.71 & 50.07  \\
    Ctx-Aware~\cite{yao:2019:docred}$^*$ & 48.40 & 50.70  \\
    BiLSTM~\cite{yao:2019:docred} & 48.78 & 51.06  \\
    Two-Step~\cite{wang:2019:two-step-bert}$^*$ & {{-}} & 53.92  \\
    HIN~\cite{Tang:2020:hin}$^*$ & 53.70 & 55.60  \\
    
    \B{\name{} (GRC)}$^*$ & 53.76 & 55.91 \\
                   
    LSR~\cite{nan:2020:bert_lsr}$^*$ & 56.97 & 59.05  \\
    CorefRo~\cite{ye:2020:coref_bert}$^*$ & 57.90 & 60.25  \\
    
    \B{\name{} (MRC)}$^*$ & \B{58.44} & \B{60.40} \\
     \bottomrule
\end{tabular}
\caption{Comparison of our relation classification component (GRC/MRC) with the state-of-the-art on the DocRED relation extraction task. We report test set results on the original DocRED split. Ign F1 ignores relational facts also present in the train set. Models marked with $*$ use a Transformer-type model for document encoding.} 
\label{table:state_art} 
\end{table}

\name{} is trained and evaluated on the end-to-end dataset split (see Table~\ref{table:joint_split}). We perform 5 runs for each experiment and report the averaged results. To study the effects of joint training, we experiment with two approaches: (a) All four sub-components are trained jointly in a single model as described in Section~\ref{sec:training} and (b) we construct a pipeline system by training each task separately and not sharing the document encoder. 

Table~\ref{table:joint_results} illustrates the results for the joint (left) and pipeline (right) approach. As described in Section~\ref{sec:approach}, each sub-task builds on the results of the previous component during inference. We observe the biggest performance drop for the relation classification task, underlining the difficulty in detecting document-level relations. Furthermore, the multi-instance based relation classifier (MRC) outperforms the global relation classifier (GRC) by about 2.4\% F1 score. We reason that the fusion of local evidences by multi-instance learning helps the model to focus on appropriate document sections and alleviates the impact of noise in long documents. Moreover, we found the multi-instance selection to offer good interpretability, usually selecting the most relevant instances (see Figure~\ref{fig:multi_instance_example} for examples). Overall, we observe a comparable performance by joint training versus using the pipeline system.

\npdecimalsign{.}
\nprounddigits{2}
\begin{table}
\sisetup{round-mode=places,detect-weight}
\centering
\begin{tabular}{l S S }
\toprule
      & \multicolumn{1}{c}{{\textbf{JM}$^*$}} & \multicolumn{1}{c}{{\textbf{SM}}} \\ \cmidrule(lr){2-2} \cmidrule(lr){3-3}
     \textbf{Task} & {F1} & {F1} \\ \midrule
     Mention Localization & 92.99438038342063 & 92.66362618180385 \\
     Coreference Resolution & 90.53607011458206 & 90.45661398542912 \\
     Entity Classification & 95.6613211711049 & 95.29056936883919 \\
     Relation Classification & 59.45818471053916 & 59.76466698664397 \\
     Relation Classification (GRC) & 56.447481187905815 & 56.5456223978627 \\
     \bottomrule
\end{tabular}
\caption{Single-task performance of the joint model (left) and separate models (right) on the end-to-end split ($^*$ joint results are for MRC except for the last row).} 
\label{table:classify_results} 
\end{table}

This is also confirmed by the results reported in Table~\ref{table:classify_results}, where we evaluate the four components independently, i.e. each component receives ground truth samples from the previous step in the hierarchy (e.g. ground truth mentions for coreference resolution). Again, we observe the performance difference between the joint and pipeline model to be negligible.
This shows that it is not necessary to build separate models for each task, which would result in training and inference overhead due to multiple expensive BERT passes. Instead, a single neural model is able to jointly learn all tasks necessary for document-level relation extraction, therefore easing training, inference and maintenance.

\begin{figure*} 
\centering
\framebox{
\begin{tabular}{ p{15cm} }
\small
    \fcolorbox{darkblue}{docgrey}{\textcolor{darkblue}{\textbf{Queequeg}}} is a fictional character in the 1851 novel Moby-Dick by American author {}\fcolorbox{darkblue}{docgrey}{\textcolor{darkblue}{\textbf{Herman Melville}}}. The son of a South Sea chieftain who left home to explore the world, \textcolor{darkblue}{\textbf{Queequeg}} is the first principal character encountered by the narrator, Ishmael. The quick friendship and relationship of equality between the tattooed cannibal and the white sailor shows \textcolor{darkblue}{\textbf{Melville}}'s basic theme of shipboard democracy and racial diversity...
    \vspace{0.1cm}
    \hrule
    \vspace{0.1cm}
    Shadowrun:Hong Kong is a turn-based tactical role-playing video game set in the Shadowrun universe. It was developed and published by \fcolorbox{darkblue}{docgrey}{\textcolor{darkblue}{\textbf{Harebrained Schemes}}}, who previously developed \fcolorbox{darkblue}{docgrey}{\textcolor{darkblue}{\textbf{Shadowrun Returns}}} and its standalone expansion. It includes a new single - player campaign and also shipped with a level editor that lets players create their own Shadowrun campaigns and share them with other players. In January 2015, \textcolor{darkblue}{\textbf{Harebrained Schemes}} launched a Kickstarter campaign in order to fund additional features and content they wanted to add to the game, but determined would not have been possible with their current budget. The initial funding goal of US \$ 100,000 was met in only a few hours. The campaign ended the following month, receiving over \$ 1.2 million. The game was developed with an improved version of the engine used with \textcolor{darkblue}{\textbf{Shadowrun Returns}} and Dragonfall. \textcolor{darkblue}{\textbf{Harebrained Schemes}} decided to develop the game only for Microsoft Windows, OS X, and Linux, ...
\end{tabular}
}
\caption{Two example documents of the DocRED dataset. Highlighted are relations \enquote{creator} between \enquote{Queequeg} and \enquote{Herman Melville} (top) and \enquote{developer} between \enquote{Shadowrun Returns} and \enquote{Harebrained Schemes} (bottom). Bordered pairs are the top selections of the multi-instance relation classifier.}
\label{fig:multi_instance_example} 
\end{figure*}

\subsection{Relation Extraction}
We also compare our model with the state-of-the-art on DocRED's relation extraction task. Here, entity clusters are assumed to be given. We train and test our relation classification component on the original DocRED dataset split. Since test set labels are hidden, we submit the best out of 5 runs on the development set via CodaLab to retrieve the test set results. Table~\ref{table:state_art} includes previously reported results from current state-of-the-art models. Note that our global classifier (GRC) is similar to the baseline by \cite{yao:2019:docred}. However, we replace mention span averaging with max-pooling and also choose max-pooling to aggregate mentions into an entity representation, yielding considerable improvement over the baseline. Using the multi-instance classifier (MRC) instead further improves performance by about 4.5\%. Here our model also outperforms complex methods based on graph attention networks~\cite{nan:2020:bert_lsr} or specialized pre-training~\cite{ye:2020:coref_bert}, achieving a new state-of-the-art result on DocRED's relation extraction task.

\subsection{Ablation Studies}

We perform several ablation studies to evaluate the contributions of our proposed multi-instance relation classifier enhancements: We remove either the global entity representations $\mathbf{x}_1^e,\mathbf{x}_2^e$ (Equation \ref{eq:entityrepr}) (a) or the localized context representation $\mathbf{c}(s_1,s_2)$ (Equation \ref{eq:mentioninput2}) (b). The performance drops by about $0.66\%$ F1 score when global entity representations are omitted, indicating that multi-instance reasoning benefits from the incorporation of entity-level context. When the localized context representation is omitted, performance is reduced by about $0.90\%$, confirming the importance of guiding the model to relevant input sections. Finally, we limit the model to fusing only intra-sentence mention pairs (c). In case no such instance exists for an entity pair, the closest (in token distance) mention pair is selected. Obviously, this modification reduces computational complexity and memory consumption, especially for large documents. Nevertheless, while we observe intra-sentence pairs to cover most relevant signals, exhaustively pairing all mentions of an entity pair yields an improvement of $0.67\%$. 

\npdecimalsign{.}
\nprounddigits{2}
\begin{table}
\sisetup{round-mode=places,detect-weight}
\centering
\begin{tabular}{l S}
\toprule
    \textbf{Model} & {F1} \\ \midrule
    Relation Classification (MRC) & 59.76466698664397 \\
    - (a) Entity Representations & 59.095287041999505 \\
    - (b) Localized Context & 58.853425842356955 \\
    - (c) Exhaustive Pairing & 59.08573665533463 \\
     \bottomrule
\end{tabular}
\caption{Ablation studies for the multi-level relation classifier (MRC) using the end-to-end split. We either remove global entity representations (a), the localized context (b) or only use intra-sentence mention pairs (c). The results are averaged over 5 runs.} 
\label{table:ablations} 
\end{table}

\section{Conclusions}
We have introduced \name{}, a novel multi-task model for end-to-end relation extraction. In contrast to prior systems, \name{} combines entity mention localization with coreference resolution to extract entity types and relations on an entity level. We report first results for entity-level, end-to-end, relation extraction as a reference for future work. Furthermore, we achieve state-of-the-art results on the DocRED relation extraction task by enhancing multi-instance reasoning with global entity representations and a localized context, outperforming several more complex solutions. We showed that training a single model jointly on all sub-tasks instead of using a pipeline approach performs roughly on par, eliminating the need of training separate models and accelerating inference. One of the remaining shortcomings lies in the detection of false positive relations, which may be expressed according to the entities' types but are actually not expressed in the document. Exploring options to reduce these false positive predictions seems to be an interesting challenge for future work.

\section*{Acknowledgments}
This work was funded by German Federal Ministry of Education and Research (Program FHprofUnt, Project DeepCA (13FH011PX6)).

\bibliography{eacl2021}
\bibliographystyle{acl_natbib}

\end{document}